\pdfoutput=1

\ifdefined\REVIEW
    \documentclass[
        a4paper,
    ]{article}
    \usepackage{lineno, setspace}
\else
    \documentclass[
        a4paper,
        twocolumn
    ]{article}
\fi

\usepackage{IEK10} \usepackage{natbib}

\usepackage{todonotes} \makeatletter
\renewcommand{\todo}[2][]{
  \tikzexternaldisable\@todo[#1]{#2}\tikzexternalenable
}
\makeatother

\def\ICE{
  Forschungszentrum Jülich GmbH,
  Institute of Climate and Energy Systems,
  Energy Systems Engineering (ICE-1),
  Jülich 52425,
  Germany
}
\def\JARA{
  JARA-CSD,
  Jülich 52425,
  Germany
}
\def\SVT{
  RWTH Aachen University,
  Process Systems Engineering (AVT.SVT),
  Aachen 52074,
  Germany
}
\def\UCL{
    University College London, 
    Sargent Centre for Process Systems Engineering,
    Department of Chemical Engineering, 
    London WC1E 7JE, 
    United Kingdom
}

\def\Wisc{
    University of Wisconsin–Madison,
    Department of Chemical \& Biological Engineering,
    Madison, WI 53706,
    USA
}
\newcommand{\mytitle}{
    Bayesian Optimization for Partially Known Systems using Hybrid Models
}

\newcommand{\affil}{
  \begin{itemize}[leftmargin=3mm, itemsep=0mm]
    \item[$^a$]\UCL
    \item[$^b$]\SVT
    \item[$^c$]\Wisc
    \item[$^d$]\JARA
    \item[$^e$]\ICE
  \end{itemize}
}

\def\firstAuthor{Eike Cramer}
\newcommand{\myauthor}{
\firstAuthor$^{a,*}$\orcidlink{0000-0002-6882-5469}
Luis Kutschat$^{b}$, 
Oliver Stollenwerk$^{b}$,
Joel A. Paulson$^{c}$\orcidlink{0000-0002-1518-7985}, 
Alexander Mitsos$^{d,b,e}\orcidlink{0000-0003-0335-6566}$
}
\newcommand{\correspondingauthor}{E. Cramer, \UCL \newline E-mail: e.cramer@ucl.ac.uk}
\author{\myauthor}

\usepackage[
  colorlinks,
  linkcolor=blue,
  citecolor=blue,
  urlcolor=blue,
  pdftitle={\mytitle},
  pdfauthor={\firstAuthor}
]{hyperref}
\usepackage[capitalise, nameinlink]{cleveref}
\crefname{table}{Tab.}{Tab.}

\usepackage{orcidlink}

\newcommand{\setpgfexternalcounter}[1]{
  \makeatletter \pgfkeysgetvalue{/tikz/external/figure name}\myexternalname
  \expandafter\gdef\csname c@tikzext@no@\myexternalname\endcsname{#1}\makeatother
}

\begin{document}

\ifx\REVIEW\undefined
\twocolumn[
\begin{@twocolumnfalse}
\fi
  \thispagestyle{firststyle}

  \begin{center}
    \begin{large}
      \textbf{\mytitle}
    \end{large} \\
    \myauthor
  \end{center}

  \vspace{0.5cm}

  \begin{footnotesize}
    \affil
  \end{footnotesize}

  \vspace{0.5cm}

\textbf{Abstract: }

Bayesian optimization (BO) has gained attention as an efficient algorithm for black-box optimization of expensive-to-evaluate systems, where the BO algorithm iteratively queries the system and suggests new trials based on a probabilistic model fitted to previous samples. 
Still, the standard BO loop may require a prohibitively large number of experiments to converge to the optimum, especially for high-dimensional and nonlinear systems. 
We present a hybrid model-based BO formulation that combines the iterative Bayesian learning of BO with partially known mechanistic physical models. 
Instead of learning a direct mapping from inputs to the objective, we write all known equations for a physics-based model and infer expressions for variables missing equations using a probabilistic model, in our case, a Gaussian process (GP). 
The final formulation then includes the GP as a constraint in the hybrid model, thereby allowing other physics-based nonlinear and implicit model constraints.
This hybrid model formulation yields a constrained, nonlinear stochastic program, which we discretize using the sample-average approximation.
In an in-silico optimization of a single-stage distillation, the hybrid BO model based on mass conservation laws yields significantly better designs than a standard BO loop.
Furthermore, the hybrid model converges in as few as one iteration, depending on the initial samples, whereas, the standard BO does not converge within 25 for any of the seeds. 
Overall, the proposed hybrid BO scheme presents a promising optimization method for partially known systems, leveraging the strengths of both mechanistic modeling and data-driven optimization. 
 \vspace*{0.5cm}

   \vspace*{5mm}
\ifx\REVIEW\undefined
\end{@twocolumnfalse}
]
\fi

\section{Introduction}
Bayesian optimization (BO)~\citep{garnett_bayesoptbook_2023} has gained attention as an efficient decision-making algorithm for optimization of black-box and expensive-to-evaluate systems, e.g., in software development and deployment~\citep{Shahriari2016Review_BO}, automated chemical discovery~\citep{Terayama2021blackboxOpt}, and robotics~\citep{Berkenkamp2021BOsafeRobotics}. 
BO solves optimization problems in a black-box setting, where the algorithm iteratively queries the system to obtain information and to converge to the optimum. 
Using Bayesian principles, BO relies on probabilistic machine learning models, typically Gaussian processes (GPs), to update its current belief about the system~\citep{garnett_bayesoptbook_2023}. 
Additional query conditions are selected by optimizing over so-called acquisition functions, which aim to balance between exploration and exploitation based on the current belief. 
The results reported in the literature indicate that BO outperforms domain experts in process optimization via experiments~\citep{Reker2020BO_outperforms_Researchers}, and that BO further leads to improved results over other design of experiments (DoE) methods~\citep{Shields2021BO_reaction}.
Examples of engineering applications of BO include the design of electrode materials~\citep{Misner2023multi_fidelity_asychronous_batch}, chemical reaction optimization~\citep{Reker2020BO_outperforms_Researchers, Shields2021BO_reaction}, controller tuning~\citep{Petsagkourakis2022CC_Policy, Lima2025InnovationsProcessControl}, and sustainable process systems~\citep{PaulsonTsay2025BO_Opinion}.

The standard BO algorithm considers all systems as black-box. However, decision makers typically have some information, e.g., through experience by an operator, lower-fidelity models, or prior experiments in parts of the system.
Possible options to include this information are prior mean modeling~\citep{Lapkin2023pHbot}, multi-fidelity BO~\citep{LeGratiet2014recursivecokriging}, and considering inequality constraints, either described by mechanistic expression~\citep{Paulson2022COBALT} or learned via additional GPs~\citep{Gardner2014BO_ineq}.
Multi-fidelity BO was shown to be very effective in incorporating multiple sources of information, such as different measurements~\citep{Misner2023multi_fidelity_asychronous_batch}, but our prior work has also shown limitations, e.g., in convergence to local optima~\citep{Cramer2025Stefans_Multifidelity_ESCAPE}. 

For physical systems such as in engineering design optimization, the available information is often embedded in incomplete mechanistic models derived from conservation laws such as Newton's laws or mass and energy balances. 
While these incomplete models cannot be used for model-based optimization alone, they can be combined with data-driven models to form so-called hybrid models. 
For non BO cases, hybrid models combining data-driven and mechanistic components have shown good performance for optimization~\citep{vonStosch2014hybrid_modeling_PSE, Schweidtmann2024hybrid_modeling_Perspective}.
However, most hybrid model formulations are not probabilistic, making them unsuitable for BO. 
Bayesian Hybrid Models (BHM) give probabilistic predictions, but are mostly used for model calibration or to formulate optimization problems without updates~\citep{Dowling2022Bayesian_hybrid_models_escape, Dowling2023Bayesian_hybrid_models}. 

\cite{astudillo2019_BO_composite} proposed the seminal idea of BO with composite functions, using the probabilistic model as the inner function and mechanistic equations in the outer equations. 
Their approach reduced the unknown component of the optimization problem to only the inner function, reducing the learning task for the probabilistic model and simplifying the optimization. 
González and Zavala build on the composite function idea using linearization to propagate the probabilistic model through the nonlinear outer equations~\citep{Gonzlez2024BOIS, Gonzlez2025implementation_BOIS}.
In related work, \cite{astudillo2021bayesianFunctionNetworks} published their work on optimization of function networks, allowing them to exploit intermediate evaluations and run BO based on a directed, acyclic graph of probabilistic models. 
They later extended their work to allow for partial evaluations~\citep{Frazier2024BOFunNetworksPartialEval}.
In recent work, we extend the BO of function networks to cyclic graphs~\citep{Paulson2026BONSAI}, relying on fixed point solvers to solve the cycle constraints, which means they have to rely on derivative-free optimization to solve the acquisition function.

In this work, we merge the concept of hybrid modeling with BO. 
Similar to the BO with composite function by \cite{astudillo2019_BO_composite}, we combine the probabilistic data-driven GP with mechanistic model equations.
In particular, we consider the case where an incomplete or idealized mechanistic model exists but one or more equations are missing to describe the true behavior of the system. 
Our approach is to first write all known mechanistic equations and only represent the unknown equations using probabilistic models, specifically GPs. 
The number of equations modeled by the GPs equal the number of variables in the mechanistic model that have no constituting equation. 
For example, in a model using Newton's first law, where the forces applied to an object are unknown, the equations describing these forces would be written via a GP. 
Opposed to the standard BO case, our hybrid model BO uses the surrogate to describe equality constraints instead of the objective function.
Unlike the BO of composite functions by \cite{astudillo2019_BO_composite}, we build on implicit model equations. 
Thus, we obtain a constrained nonlinear program (NLP) where some of the equality constraints are mechanistic equations, and others are GPs. 
The constrained NLP formulation presents a key distinction to previous BO approaches and opens the possibility to include a multitude of different physics-based constraints, introduce new variables, and utilize established NLP paradigms. 

Writing GPs as equality constraints in an NLP introduces two challenges. 
First, the output of a GP is a random variable, making the NLP a stochastic program.
Second, the acquisition function needs to be defined based on the mechanistic objective function, and cannot be formulated using the moments of the GP as typically done in BO. 
We formulate the NLP with embedded GPs as a stochastic program~\citep{Ruszczyski2021TwoStageProblems}, where the BO decision variables are the degrees of freedom variables and the remaining states are dependent variables. 
Next, we discretize the stochastic program via sample average approximation (SAA)~\citep{Shapiro2009SAA}, by sampling from the GP distribution, turning the stochastic program into a deterministic problem. 
Furthermore, we use the discretized stochastic program to derive a formulation of the expected improvement (EI) acquisition function, which we call the acquisition problem. 
We implement the hybrid model BO and the GPs using the automatic differentiation library CasADi~\citep{CasADi} and solve the SAA discretized NLP using the nonlinear solver IPOPT~\citep{ipopt}. 

We include two case studies: an illustrative univariate case study and a chemical engineering example, where we consider the optimization of a flash unit. 
For the flash unit, we can write mass balances and ideal thermodynamics using mechanistic expressions. 
The real, nonideal thermodynamics are unknown and therefore modeled using a GP.
In both case studies, the hybrid model BO demonstrates promising performance. 

In summary, the contribution of this work is twofold.
First, we present a stochastic programming formulation for acquisition problems based on mechanistic models in combination with GPs. 
Second, we provide a Python-based modeling library for GP training and formulation of NLPs with GPs embedded using the same model instance. 

The remainder of this work is organized as follows:
Section~\ref{sec: Preliminaries Bayesian Optimization} reviews the most important concepts about BO, and the knowledgeable reader may skip ahead to later sections. 
Section~\ref{sec: Contribution BO with hybrid} introduces our approach for BO with hybrid models, including embeddings of GPs in NLPs and formulating the acquisition problem, as well as discusses how to obtain the necessary data to fit the embedded GP. 
Section~\ref{sec: Implementation} introduces our implementation using CasADi~\citep{CasADi} and solution using IPOPT~\citep{ipopt}. 
Next, Section~\ref{sec: illustrative example} applies our approach to an illustrative optimization case study and draws comparisons to benchmarks in standard BO and random uniform sampling. 
Section~\ref{sec: flash unit} considers the optimization of a flash unit with unknown thermodynamics. 
Finally, Section~\ref{sec: conclusion} concludes this work and points to further extensions.

\section{Bayesian Optimization}\label{sec: Preliminaries Bayesian Optimization}
The baseline application for BO is the optimization of black-box and expensive-to-evaluate functions such as high fidelity simulations or physical experiments~\citep{garnett_bayesoptbook_2023}. 
Hence, BO considers the case of an objective function $\phi$ that has no closed-form expression, and no gradient information is available.
Throughout this work, we restrict the analysis to scalar objectives, where the optimization problem reads as follows:
\begin{equation}\label{prob: generic black box optimization problem}
    \begin{aligned}
    \underset{\mathbf{u}}{\min}&~\phi(\mathbf{u}) \\
    \text{s.t.} &~ \mathbf{u}_{\mathrm{LB}}\leq \mathbf{u} \leq \mathbf{u}_{\mathrm{UB}}
    \end{aligned}
\end{equation}
Here, $\mathbf{u}\in \mathbb{R}^N$ are the decision variables of the optimization with lower and upper bounds $\mathbf{u}_{\mathrm{LB}}$ and $\mathbf{u}_{\mathrm{UB}}$, i.e., the variables the user may actively manipulate to minimize $\phi$. 
Note that, depending on the field, the variables in $\mathbf{u}$ are sometimes called degrees of freedom (DoF) or parameters. 
In what follows, we use the term decision variables to denote the variables that we can manipulate in the optimization. 
An important note is that the host set for the decision variables $\mathcal{U}=\{ \mathbf{u} \mid \mathbf{u}_{\mathrm{LB}} \leq \mathbf{u} \leq \mathbf{u}_{\mathrm{UB}} \}$ may include infeasible decisions, as the feasible set is unknown in BO. 
For such infeasible decisions, the unknown function $\phi$ returns no value.

For BO, the unknown objective $\phi$ is approximated via a probabilistic data-driven model, typically a GP~\citep{Rasmussen2005_GP4ML}:
\begin{equation}\label{eq: general unknown objective approx}
    \phi(\mathbf{u}) \approx \mathcal{GP}_{\phi}(\mathbf{u}\vert \mathcal{D}_{\mathbf{u},\phi})
\end{equation}
Here, $\mathcal{D}_{\mathbf{u},\phi}=\{\mathbf{u}_i,\phi_i\}_{i=1,\dots, N}$ is the dataset of trial conditions, i.e., values for decision variables $\mathbf{u}_i$, and the corresponding evaluations of the objective $\phi_i$. 
GPs have been studied extensively in the literature on BO and beyond. 
For brevity, we refer to \cite{Rasmussen2005_GP4ML} and \cite{garnett_bayesoptbook_2023} for rigorous introductions to GPs.
In the following, we omit the conditioning on $\mathcal{D}$.

BO is carried out in a loop format. The loop consists of computing optimal trial conditions $\mathbf{u}_i^*$, obtaining data for these trial conditions, and updating the model based on the new information. 
The trial conditions $\mathbf{u}_i^*$ for the i-th iteration of the loop can be computed by inserting the mean function of the GP approximation to Problem~\eqref{prob: generic black box optimization problem}, which corresponds to a greedy search. 
Alternatively, we can formulate so-called acquisition functions aiming to balance between exploration and exploitation of the system given the currently available information~\citep{garnett_bayesoptbook_2023}.

There are many acquisition functions for BO~\citep{garnett_bayesoptbook_2023}, but throughout this work, we focus on the Expected Improvement (EI). The EI acquisition function describes the improvement in either maximization or minimization over an incumbent $\phi'$, a constant that represents the currently best known sample of the system.
The EI is most often defined for maximization problems. In this work, we consider minimization problems leading to the following reformulation: 
\begin{equation}\label{eq: EI general minimize}
    \alpha_{\mathrm{EI}}(\mathbf{u}) = \mathbb{E}_{\phi}\left[\left(\mathcal{GP}_{\phi}(\mathbf{u}) - \phi'\right)^{-} \right]
\end{equation}
Here, $\mathcal{GP}_{\phi}$ is the GP approximation of the true objective $\phi$, and the $(\cdot)^-$ operator indicates $\min(0,\cdot)$.
In Appendix~\ref{app: LCB} we also discuss the lower confidence bound (LCB).
The EI acquisition function includes the expectation operator because the GP posterior is a random variable. 
For the standard BO setting, there exists an analytical reformulation of the EI based on the GP approximation~\citep{Rasmussen2005_GP4ML, garnett_bayesoptbook_2023}.

\section{Bayesian Optimization with Hybrid Models}
\label{sec: Contribution BO with hybrid}
This section derives our approach to incorporate mechanistic models into the BO framework and derives the stochastic program we call the acquisition problem.

\subsection{Nonlinear Programs with Gaussian Processes Embedded}\label{ssec: NLP with GP}

For the introduction of our approach to BO with hybrid models, we first define a general version of an optimization problem:
\begin{equation}\label{prob: generic optimization problem}
    \begin{aligned} 
        \underset{\mathbf{x},\mathbf{y},\mathbf{u}}{\min} ~& f(\mathbf{x},\mathbf{y},\mathbf{u}) \\
        \text{s.t.} ~ 
        \mathbf{0} &= \mathbf{g}(\mathbf{x},\mathbf{y},\mathbf{u})  \\
        \mathbf{0} &= \mathbf{h}(\mathbf{x},\mathbf{y},\mathbf{u})  \\
\mathbf{u}_{\mathrm{LB}}&\leq \mathbf{u} \leq \mathbf{u}_{\mathrm{UB}} \\
    \end{aligned}
\end{equation}
In this problem formulation, $\mathbf{u}\in\mathbb{R}^N$ are decision variables bounded by their lower and upper bounds, respectively. The variables $\mathbf{x}\in\mathbb{R}^M$ and $\mathbf{y}\in\mathbb{R}^P$ are dependent variables that are determined via the value of $\mathbf{u}$ and the model equations. 
The dependent variables $\mathbf{x}$ and $\mathbf{y}$ may have variable bounds, but these are not necessary for the problem formulation.
The function $f$ is the known objective, $\mathbf{g}$ are implicit functions describing equality constraints known in a mechanistic form, and the function $\mathbf{h}$ represents the unknown equations.
We split the dependent variables into $\mathbf{x}$ and $\mathbf{y}$, where 
$\dim \mathbf{g}=\dim \mathbf{x}$ and
$\dim \mathbf{h}=\dim \mathbf{y}$, to isolate the dependencies of the mechanistic equations $f$ and $\mathbf{g}$ from the unknown model components, which we summarize in the unknown function $\mathbf{h}$. 
Problem~\eqref{prob: generic optimization problem} may include inequality constraints as well, which we omit here for the sake of brevity. 
Note that for $\dim \mathbf{y}=1$, $f(\mathbf{x},\mathbf{y},\mathbf{u})=\mathbf{y}$, and no equality constraints $\mathbf{g}$, Problem~\eqref{prob: generic optimization problem} is equivalent to the standard fully black-box BO problem, see Problem~\eqref{prob: generic black box optimization problem}.

Since we have mechanistic expressions for $f$ and $\mathbf{g}$, there is no need to approximate the objective $f$ using the GP as one would do in the baseline BO setting. 
Instead, we use the GP to approximate only the unknown equations $\mathbf{h}$:
\begin{equation}\label{eq: approximation of unknown h with GP}
    \mathbf{y}\approx \mathcal{GP}_{\mathbf{h}}(\mathbf{x},\mathbf{u}) 
\end{equation}

Modeling the unknown parts as a GP makes $\mathbf{h}$ a probabilistic function and the variable $\mathbf{y}$ a random variable. 
We formulate the GP using the reparameterization trick~\citep{Kingma2013VariationalBayesAndReparameterization}:
\begin{equation}\label{eq: reparameterization trick}
    \mathcal{GP}_{\mathbf{h}}(\mathbf{x},\mathbf{u}) = \bm{\mu}_{\mathbf{y}}(\mathbf{x},\mathbf{u}) + \mathbf{L}_{\mathbf{y}} (\mathbf{x},\mathbf{u})\bm{\xi}
\end{equation}
Here, $\bm{\mu}_{\mathbf{y}}(\mathbf{x},\mathbf{u})$ is the posterior mean and $\mathbf{L}_{\mathbf{y}} (\mathbf{x},\mathbf{u})$ is the lower triangular matrix of the Cholesky decomposition of the covariance matrix of $\mathcal{GP}_{h}(\mathbf{x},\mathbf{u})$.
Both $\bm{\mu}_{\mathbf{y}}(\mathbf{x},\mathbf{u})$ and $\mathbf{L}_{\mathbf{y}}(\mathbf{x},\mathbf{u})$ are deterministic functions, and the random variable $\xi$ is a standard normal with $\bm{\xi}\sim\mathcal{N}(\mathbf{0},\mathbf{I})$ and $\dim\bm{\xi}=\dim\mathbf{y}$.
By writing the GP in terms of the reparameterization, Equation~\eqref{eq: reparameterization trick} isolates the random variable $\bm{\xi}$ without any dependence on other variables, and we can implement $\bm{\mu}_{\mathbf{y}}(\mathbf{x},\mathbf{u})$ and $\mathbf{L}_{\mathbf{y}}(\mathbf{x},\mathbf{u})$ as deterministic functions in the optimization problem.

Entering the GP approximation in Equation~\eqref{eq: reparameterization trick} for the unknown equation $\mathbf{h}$ in Problem~\eqref{prob: generic optimization problem}, results in a stochastic program, where we minimize the expectation over the objective. 
\begin{equation}\label{prob: stochastic program with GP}
    \begin{aligned} 
        \underset{\mathbf{u}}{\min} ~& \mathbb{E}_{\bm{\xi}}\left[f(\mathbf{x},\mathbf{y},\mathbf{u})\right] \\
        \text{s.t.}  
~ \mathbf{0} &= \mathbf{g}(\mathbf{x},\mathbf{y},\mathbf{u})  \\
         \mathbf{y} &=\bm{\mu}_{\mathbf{y}}(\mathbf{x},\mathbf{u}) + \mathbf{L}_{\mathbf{y}} (\mathbf{x},\mathbf{u}) \bm{\xi} \\
         \mathbf{u}_{lb}&\leq \mathbf{u} \leq \mathbf{u}^{ub} \\
    \end{aligned}
\end{equation}
Here, the dependent variables $\mathbf{x}$ and $\mathbf{y}$ are determined based on $\mathbf{u}$ and the realization of $\bm{\xi}$.
In that sense, one could consider Problem~\eqref{prob: stochastic program with GP} a two-stage program~\citep{Ruszczyski2021TwoStageProblems}. 
However, the recourse variables are uniquely defined by the solution of the equality constraints.

We emphasize that we utilize the definitions and formulations from stochastic programming to formulate Problem~\eqref{prob: stochastic program with GP}, but there is no aleatoric uncertainty in the system, i.e., there is no stochastic process. 
Instead, the random variable introduced by the GP is a quantitative estimate of the epistemic uncertainty, i.e., the model uncertainty or simply the unknown. 
This consideration tells us that any individual realization of the random variable $\bm{\xi}$ describes a possible version of the truth rather than an outcome of a stochastic process that may change at a later time. 
For a general introduction to stochastic programming, we refer to \cite{Shapiro2021LecturesStochasticProgramming}.

Due to the nonlinear mechanistic constraints in $\mathbf{g}$, the nonlinear objectives $f$, and the nonlinear GP approximation of $\mathbf{h}$, no closed-form solution for the expectation in Problem~\eqref{prob: stochastic program with GP} exists. 
Instead, we discretize the random variable $\bm{\xi}$ using SAA~\citep{Shapiro2009SAA}. 
Here, we draw Monte Carlo samples of the standard normal $\bm{\xi}$ and write the constraints such that they should hold for every scenario $\bm{\xi}_s \ \forall s=1,\dots, S$:
\begin{equation}\label{prob: SAA stochastic program}
    \begin{aligned} 
\underset{\mathbf{x}_s,\mathbf{y}_s,\mathbf{u}}{\min} ~& \frac{1}{S}\sum_{s=1}^S f(\mathbf{x}_s,\mathbf{y}_s,\mathbf{u}) \\
        \text{s.t.}      ~ 
    \mathbf{0} &= \mathbf{g}(\mathbf{x}_s,\mathbf{y}_s,\mathbf{u})  & \forall s=1,\dots,S\\
    \mathbf{y}_s &= \bm{\mu}_{\mathbf{y}}(\mathbf{x}_s,\mathbf{u}) + \mathbf{L}_{\mathbf{y}}(\mathbf{x}_s,\mathbf{u}) \bm{\xi}_s & \forall s=1,\dots,S \\
\mathbf{u}_{\mathrm{LB}} &\leq \mathbf{u} \leq \mathbf{u}_{\mathrm{UB}} \\
    \end{aligned}
\end{equation}
Here, $S$ is the total number of scenarios. 
The discretization results in a sparse block structure with an equation block for each scenario $\bm{\xi}_s$. 
Note that the decision variables $\mathbf{u}$ are not discretized, as they must be valid for all realizations of $\bm{\xi}$.
Problem~\eqref{prob: SAA stochastic program} now constitutes a sparse constrained NLP, which can be solved using NLP paradigms~\citep{Biegler2010_NLP} and solvers such as IPOPT~\citep{ipopt}.
The SAA formulation is sometimes called the deterministic equivalent of the problem~\citep{Shapiro2009SAA}.

The stochastic program in Problem~\eqref{prob: stochastic program with GP} with the discretization in Problem~\eqref{prob: SAA stochastic program} can be used in a BO format but represent a greedy scheme, as the optimization aims to minimize the expected value of the objective. 
To balance exploration and exploitation, we need to adapt the acquisition functions defined in Section~\ref{sec: Preliminaries Bayesian Optimization} for the hybrid model case.

\subsection{Acquisition Problems}\label{ssec: ACQ function Hybrid}
In the hybrid model case, the unknown function is no longer the objective of the optimization problem. 
Still, the balance between exploitation and exploration for BO should be based on the objective.
Thus, we derive what we now call acquisition problems that are based on the stochastic programs written in Section~\ref{ssec: NLP with GP}.

For the EI, the analytical reformulation based on the moments of the GP surrogate does not apply to the hybrid BO in Problem~\eqref{prob: stochastic program with GP} or the discretized form in Problem~\eqref{prob: SAA stochastic program}. 
There are many options to reformulate the $\min(0,\cdot)$ expression in the EI function. 
For our purposes, we find that maintaining the nonsmooth structure is sufficient in practice. 
\begin{equation}\label{eq: SAA EI}
    \underset{\mathbf{x}_s,\mathbf{y}_s,\mathbf{u}}{\min} ~ \frac{1}{S}\sum_{s=1}^S 
    \left( f(\mathbf{x}_s,\mathbf{y}_s,\mathbf{u})~ 
     - f'\right)^-
\end{equation}
This reformulation requires no further modification of the acquisition problem, and we insert Equation~\eqref{eq: SAA EI} as the objective in Problem~\eqref{prob: SAA stochastic program}. 
Again, $f'$ is the incumbent from previous trials. 
Equation~\eqref{eq: SAA EI} is the same approximation used by \cite{astudillo2019_BO_composite}, who solve their composite function BO using derivative-free solvers. 
In this work, we use second-order solvers such as IPOPT~\citep{ipopt}, which require smooth problem formulations. However, we find that in practice we see no issues resulting from the formulation in Equation~\eqref{eq: SAA EI}. 
For cases where the nonsmooth acquisition problem becomes an issue, Appendix~\ref{app: smooth SAA EI} lists possible smooth reformulations of Equation~\eqref{eq: SAA EI}.
Furthermore, Appendix~\ref{app: LCB} introduces the acquisition problem for the LCB for completeness. 
In the following, we refer to Equation~\eqref{eq: SAA EI} as the SAA-EI formulation.

\subsection{Sampling in Hybrid Model BO}\label{ssec: Data for the GP}
The standard BO algorithm uses a dataset of trial conditions $\mathbf{u}_i$ and the corresponding evaluations of the objective $\phi_i$ to obtain the training dataset for the GP surrogate $\mathcal{D}_{\mathbf{u},\phi}=\{\mathbf{u}_i,\phi_i\}_{i=1,\dots, N}$, as described in Section~\ref{sec: Preliminaries Bayesian Optimization}.
This approach inherently assumes that the objective $\phi$ is either measurable or can be computed from other measurements. 
For the hybrid model BO, the GP training data differ from the standard case, as the GP describes a model equation instead of the objective.

We write the GP as an equality constraint of an NLP to get an expression for the unknown function $\mathbf{0}=\mathbf{h}(\mathbf{x}, \mathbf{y}, \mathbf{u})$, and the labels for the surrogate are the variables $\mathbf{y}$, see Equation~\eqref{eq: approximation of unknown h with GP}.
The variables $\mathbf{x}$ and $\mathbf{y}$ are dependent variables and in the physical system fully determined by the value of the decision variable $\mathbf{u}$.
Thus, we can always write the GP approximation in the following form: 
\begin{equation*}
    \mathbf{y}\approx\mathcal{GP}_h(\mathbf{u})
\end{equation*}
The GP training data for this case is simply given by $\mathcal{D}_{\mathbf{u},\mathbf{y}}=\{\mathbf{u}_i,\mathbf{y}_i\}_{i=1,\dots, N}$. 
Note that, as opposed to the standard scalar objective BO case, the labels $\mathbf{y}$ may be multivariate. 

In the more general case written in Equation~\eqref{eq: approximation of unknown h with GP}, $\mathbf{y}$ depends on both the decision variables $\mathbf{u}$ and the other states $\mathbf{x}$. 
The reason to consider $\mathbf{x}$ as regressors in the GP may be given by existing physical knowledge that this specific functional relation between the variables $\mathbf{x}$ and $\mathbf{y}$ exists. 
Thus, the training dataset for the GP in this case is $\mathcal{D}_{[\mathbf{u},\mathbf{x}],\mathbf{y}}=\left\{[\mathbf{u}_i,\mathbf{x}_i],\mathbf{y}_i\right\}_{i=1,\dots, N}$, where $[\mathbf{u}_i,\mathbf{x}_i]$ are the inputs to the GP for the i-th trial. 
The selection of training data for the GP $\mathcal{D}_{[\mathbf{u},\mathbf{x}],\mathbf{y}}$ may include the full state vector $\mathbf{x}$ and the decision variable vector $\mathbf{u}$.
In cases where existing knowledge provides an informed selection, the smallest subset of variables in $\mathbf{x}$ and $\mathbf{u}$ should be selected to reduce the dimensionality of the GP to simplify the learning problem.

In the simplest case, the dependent variables $\mathbf{y}$ can be measured directly.
However, this will not be possible in all cases. 
Instead, we obtain the GP training data via a combination of measurements and simulation. 
At least $d=\dim \mathbf{y}$ variables from the dependent variable vectors $\mathbf{x}$ and $\mathbf{y}$ must be measurable or otherwise attainable. 
For fixed trial conditions $\mathbf{u}_i$ and the fixed corresponding $d$ measurements, we can solve the equation system of known mechanistic equations with zero DoF. 
Writing the measurable variables as $\mathbf{y}^m$ and $\mathbf{x}^m$ with $\dim \left[\mathbf{y}^m, \mathbf{x}^m\right]^T = d$ and the remaining variables as $\mathbf{y}^v$ and $\mathbf{x}^v$, the resulting root finding problem reads:
\begin{equation*}
\mathbf{g}(\mathbf{x}^m=\mathbf{x}^m_i,\mathbf{x}^v, \mathbf{y}^m=\mathbf{y}^m_i, \mathbf{y}^v,\mathbf{u}=\mathbf{u}_i) = \mathbf{0}
\end{equation*}
The solution to the root-finding problem then yields the full sample value vectors $\mathbf{x}_i$ and $\mathbf{y}_i$.

\section{Implementation}
\label{sec: Implementation}
To solve the acquisition problem, we need to implement the problem in a modeling framework and interface to a numerical solver for constrained NLP. 
There exist frameworks for embedding machine learning models into optimization problems. 
For instance, the Machine Learning models for Optimization (MeLOn) toolbox allows users to embed GPs in global optimization problems~\citep{Schweidtmann2021DGO_GP}. 
However, the discretized formulations of the stochastic program in Problem~\eqref{prob: SAA stochastic program} are high dimensional and therefore infeasible for global optimization. 
Another popular choice for hybrid models presents the modeling framework pyomo~\citep{pyomo} with its extension OMLT~\citep{omlt}. However, pyomo and OMLT currently do not support GPs. 
Hence, we implement a custom framework for GP training and embedding in NLP based on the Python version of CasADi~\citep{CasADi}. 
We opt for the implementation in CasADi to leverage the matrix algebra capabilities, including the differentiable Cholesky decomposition, which are currently not available in pyomo. 
Our framework allows the user to fit the hyperparameters of the GP using powerful second-order optimizers such as IPOPT~\citep{ipopt} and use the same GP-object instance to write the constrained acquisition problem. 
As the discretized acquisition problem is a block triangular system, IPOPT is able to exploit this sparsity for faster convergence.
To obtain $\mathbf{x}$ and $\mathbf{y}$ by solving $\mathbf{g}$ for the GP training data, we use the \texttt{fsolve} function in the Python-based library SciPy~\citep{scipy2020pythonlibrary}.

We publish our library for GP fitting and embedding in NLP via \hyperlink{https://git.rwth-aachen.de/avt-svt/public/hybrid-bo}{gitlab}. 
The library includes a variety of different options for kernels and priors for the GPs. 
To formulate the optimization problems, the mean and the Cholesky decomposition of the covariance function are written as deterministic functions, as described via the reparameterization trick in Equation~\eqref{eq: reparameterization trick}. 

Solving the acquisition problem and the root-finding problem for the GP training data requires initial guesses for the respective nonlinear solvers. We sample these initial guesses using space-filling designs, namely Latin hypercube sampling. 
As the GP approximation of the unknown equation is explicit (see Equation~\eqref{eq: approximation of unknown h with GP}), we can implement the functional form in CasADi and thereby avoid sampling initial guesses for the dependent variable $\mathbf{y}$. 
For the decision variables $\mathbf{u}$, the initial guesses are sampled within the variable bounds, and for the dependent variables $\mathbf{x}$, the sampling bounds can either be set or derived from existing variable bounds.
Since GP approximations are usually non-convex, we solve both optimization and root-finding problems using multi-starts to increase the probability of finding the global optimum.

Throughout the following analysis, we use the Matern52 Kernel and a constant zero prior mean for all examples. 
All GP fitting problems and all acquisition problems are solved using IPOPT in standard settings.

\section{Illustrative Example}\label{sec: illustrative example}
As our first case study, we consider an illustrative case study with a univariate unknown function. 
\subsection{Problem Description}
The optimization problem for the illustrative case study is the well established Forrester function~\citep{Mainini2025BenchmarkFunctions} with a modification to a constrained optimization problem: 
\begin{equation}\label{prob: illustrative example}
\begin{aligned}
    \underset{u, x, y}{\min} \ & f_F \left( x \right) \\
    \text{s.t.} \ & x + \exp \left( x \right) = y \\
    & y =  h\left( u \right) \\
    & -2.0 \leq u \leq 2.0 
\end{aligned} 
\end{equation}
Here, all equations and variables are scalars, $f_F$ is the Forrester function, and $h$ represents the unknown component of the problem. 
For this illustrative example, we have the following oracle function:
\begin{equation}\label{eq: illustrative oracle}
h(u) = \sin\left(u\right)
\end{equation}
The oracle function depends only on the decision variable $u$ without any dependence on $x$. 
Therefore, we can query the oracle and train the GP on the results without further computations, as described in Section~\ref{ssec: Data for the GP}.
This version of the general formulation allows us to visualize the propagation of the GP variable through the NLP.

To solve Problem~\eqref{prob: illustrative example}, we discretize the random normal $\xi$ using 25 scenarios $\xi_s$ and sample two initial points using Latin hypercube sampling. 
Next, we write the acquisition problem using the SAA-EI formulation in Equation~\eqref{eq: SAA EI}. 
Both the fitting of the GP hyperparameters and the optimization of the acquisition problem are run in multi-start formats with 100 initial guesses generated using Latin hypercube sampling each. 
We use IPOPT~\citep{ipopt} in standard settings to fit the GP and solve the acquisition problem. 

\subsection{Convergence}

\begin{figure}
    \centering
    \includegraphics[width=\columnwidth]{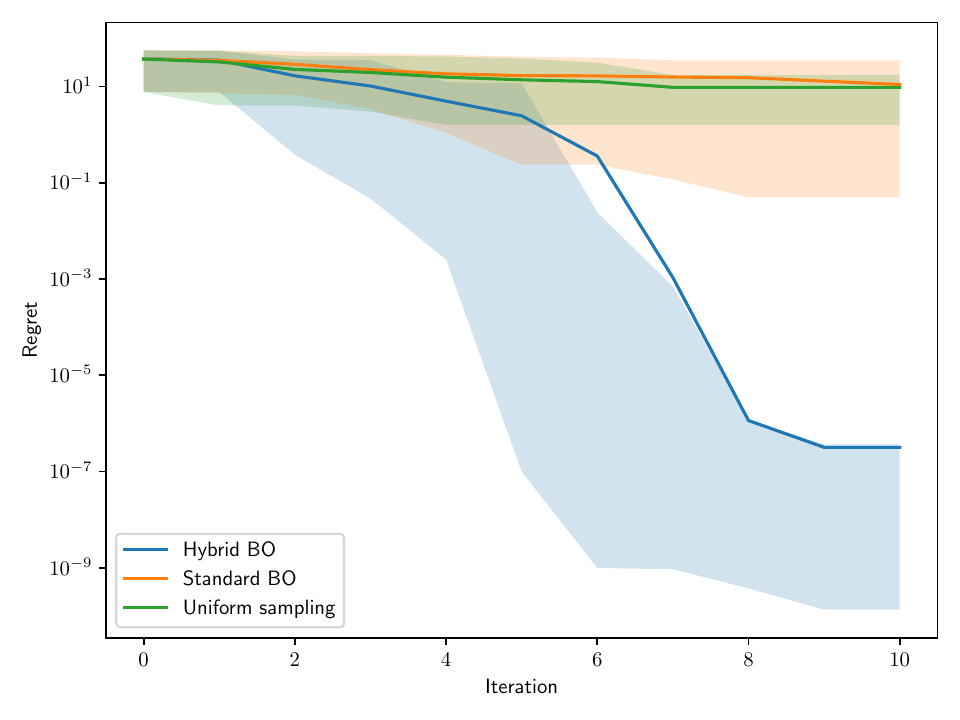}
    \caption{
        Log-regret $\mathrm{Regret} = f'-f^*$ of BO for the illustrative example for random sampling, standard BO with EI, and hybrid-BO with SAA-EI. 
        The shaded areas represent the 90\% intervals of the regret convergence over 25 runs. 
        }
    \label{fig: Regret Abstract}
\end{figure}
Figure~\ref{fig: Regret Abstract} shows the convergence of the BO loop with the hybrid model and SAA-EI in comparison to random uniform sampling and a standard BO with EI, where the objective is modeled directly via the GP.
Using a logarithmic scaling, the figure shows the regret, which is defined as follows~\citep{garnett_bayesoptbook_2023}:
\begin{align} \label{eq: simple regret} 
    \mathrm{Regret} = f'-f^*
\end{align}
Here, $f'$ denotes the current incumbent, and $f^*$ denotes the global optimum, which we can compute as we know the oracle function in Equation~\eqref{eq: illustrative oracle}.
We run each approach for 25 distinct initial sample sets to obtain statistical results across multiple runs. 

The regret progression in Figure~\ref{fig: Regret Abstract} shows similar convergence of the proposed hybrid model BO and the standard BO for the first two iterations. 
Afterwards, the hybrid model BO continues to improve significantly while the standard BO appears to converge. After ten iterations, the hybrid model BO has identified solutions with regrets seven orders of magnitude lower than the standard BO. 
The random uniform sampling performs similarly to the standard BO, which is a result of the small one-dimensional decision space, where random sampling has high probabilities of finding good results.

\subsection{Sampling Analysis}\label{ssec: illustrative sampling analysis}
\begin{figure*}
    \centering
\includegraphics[width=0.32\textwidth]{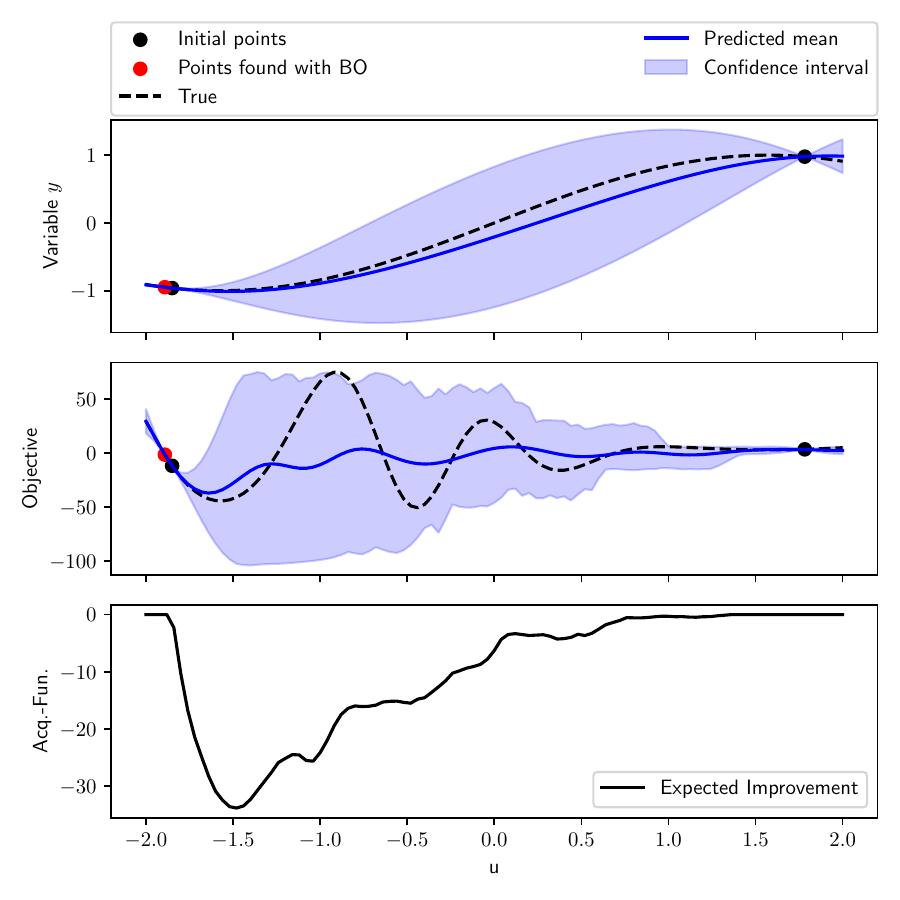}
    \includegraphics[width=0.32\textwidth]{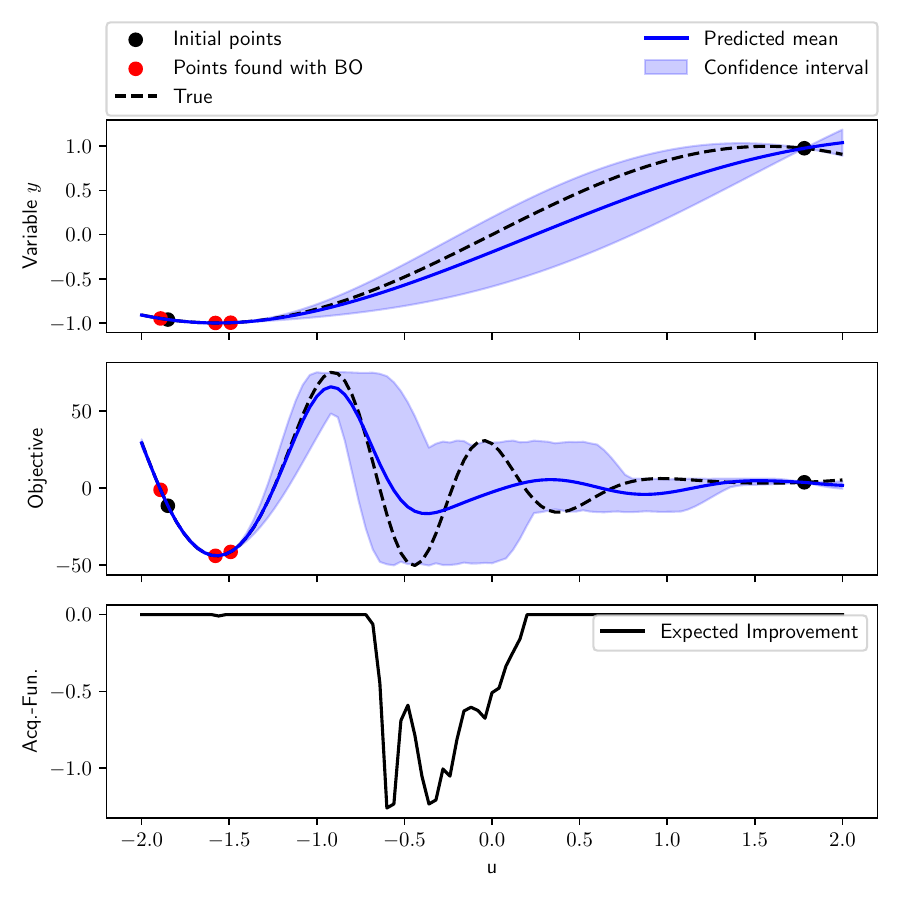}
    \includegraphics[width=0.32\textwidth]{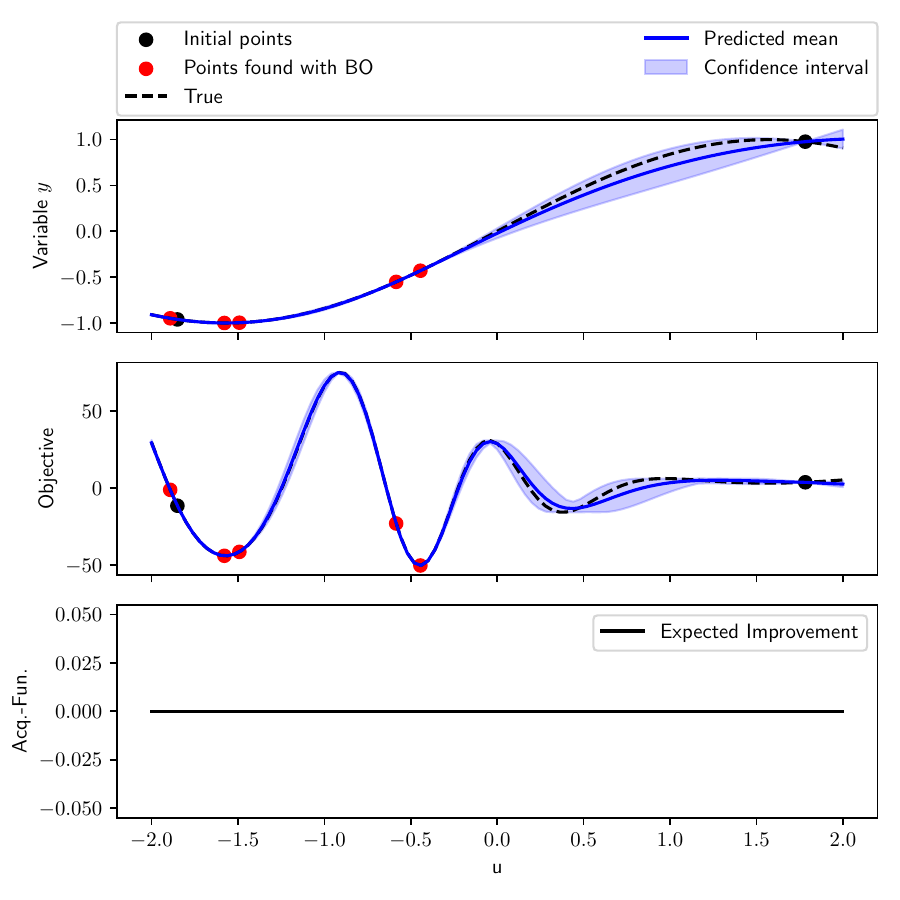}
    \caption{
    Progression of SAA-EI with hybrid acquisition problem~\eqref{prob: illustrative example}.
    The top row shows the GP approximation of the unknown equation $h(u)\approx\mathcal{GP}(u)$, the center row shows the approximation of the objective, and the bottom row shows the SAA-EI (Equation~\eqref{eq: SAA EI}). 
    The approximations show the 90\% confidence intervals, where the approximation of the objective is displayed as the quantiles estimated from 25 discrete samples $\xi_s$. 
    Left: Iteration 1; center: Iteration 3; right: Iteration 6. 
    }
    \label{fig: illustrative example approximations}
\end{figure*}
Figure~\ref{fig: illustrative example approximations} shows the approximation of the unknown function $h\approx\mathcal{GP}(u)$ in Equation~\eqref{eq: illustrative oracle}, the approximation of the objective, and the SAA-EI function for the first iteration (left), the third iteration (center), and the sixth iteration (right), respectively. 
The shaded areas indicate the 90\% confidence intervals of the GP and its propagation through the nonlinear model, i.e., the 5 and 95 quantiles estimated from the samples. 
The visualization uses the 25 scenarios for $\xi_s$ to estimate the quantiles. 

The GP approximation for the first iteration shows large confidence intervals, as there are only three total samples. The wide confidence intervals highlight the nonlinear transformation of the GP samples to compute the objective. 
Notice that the confidence intervals for the objective are asymmetrical around the mean estimate, as they result from the nonlinear transformation. 
The confidence intervals of the objective clearly show the approximation via SAA visible by their non-convex nature. The SAA approximation consequently leads to a non-convex acquisition function with many local minima further emphasizing the need for the high number of multi-starts for the aquisition problem. 
After three BO iterations in the center of Figure~\ref{fig: illustrative example approximations}, the approximation is much improved, and the absolute values of the SAA-EI have reduced by close to two orders of magnitude. 
Again, the nonlinear transformation of the GP samples remains visible and shows the approximation via SAA. 
The SAA-EI function, after the third iteration, has two minima close to the global optimum. 
After six iterations in the right of Figure~\ref{fig: illustrative example approximations}, the BO loop is converged close to the global optimum, and the SAA-EI function is zero throughout the domain. 
Note that we write that convergence is close to the global optimum as the SAA-EI function is based on a discretized approximation of the objective function, see Equation~\eqref{eq: SAA EI}. 
Hence, it is not clear whether there is some small value for the tails of the distribution of the objective and the true EI that is not recognized by the SAA-EI approximation.

\subsection{Discussion}
The analysis of the illustrative example shows that the hybrid model formulation of the acquisition problem samples the problem efficiently and converges to a near-optimal solution with very few samples, despite the challenges posed by the highly non-convex SAA-EI acquisition function. 
For the simple illustrative example, the hybrid acquisition problem converges to significantly lower regret values than the standard BO with EI.  
Overall, the illustrative example shows highly promising results for further investigation. 
Still, the highly non-convex acquisition functions resulting from the SAA-EI approximation present a challenging optimization problem. 
Furthermore, the poor approximation of the tail of the objective function distribution leads to constant zero SAA-EI values throughout the domain, limiting the optimization to convergence to near-optimal points, even for many BO iterations.

\section{Flash Unit Optimization}\label{sec: flash unit}
We now consider the optimization of a flash unit to separate water and acetic acid.
Flash units are common unit operations in chemical plants for a single separation step of liquid and vapor mixtures.

\subsection{Flash Unit Model}\label{ssec: flash unit model}
\begin{figure}
    \centering
    \includegraphics[width=\columnwidth]{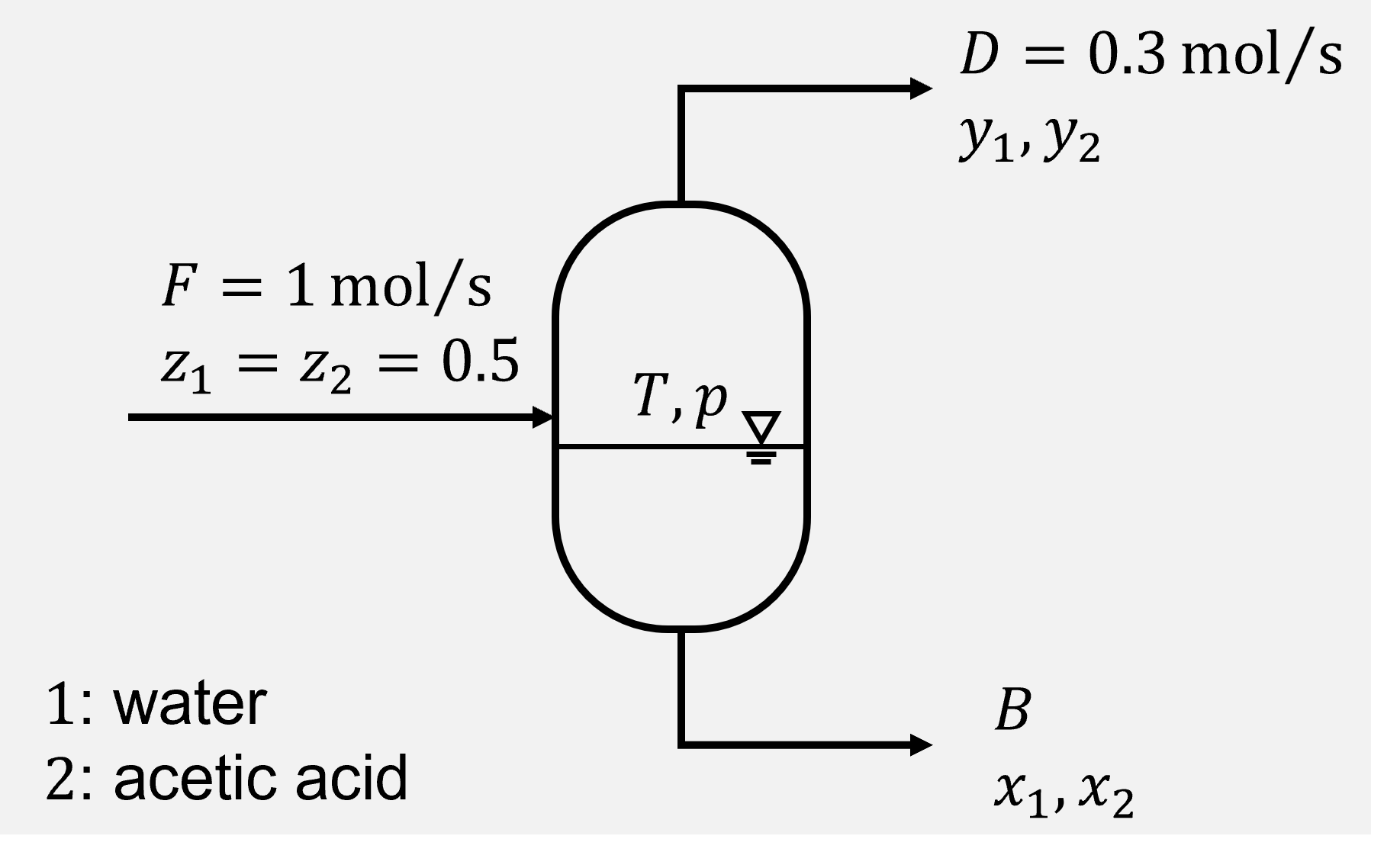}
    \caption{Sketch of the flash unit with fixed inlet stream composition and temperature $T$ and pressure $p$ as the decision variables.}
    \label{fig: sketch flash}
\end{figure}
A simple flash unit can be modeled using conservation laws and pure-component thermodynamics, with additional equations describing mixture thermodynamics. 
For nonideal mixtures such as water and acetic acid, equations describing the real thermodynamics of the mixture are difficult and expensive to obtain. 
Figure~\ref{fig: sketch flash} shows a sketch of the flash unit. 
We formulate the following objective, aiming to set a specific distillate mole fraction while minimizing the cost from heating and pressure difference to the environment: 
\begin{equation}\label{eq: flash Obj}
    \underset{T,p}{\min} ~ \zeta_1 (y_1-y^{set}_1)^2 + \zeta_2 T^2 + \zeta_3 (p-p^{atm})^2
\end{equation}
Here, we select the values $\zeta_1$, $\zeta_2$, and $\zeta_3$ as weights to balance between the different components. 
The decision variables for the flash unit are the temperature $T$ and pressure $p$ with $T\in [363.15\,\text{K}, 403.15\,\text{K}]$ and $p\in[0.8\,\text{bars}, 2.6\,\text{bars}]$. 

The flash has a known fixed feed and a specified total molar flow at the head. 
We may write the following model of the flash system:
\begin{subequations}
\label{eq: flash model}
    \begin{align}
        0 &= D+B -F \label{eq: Mass Balance}\\
        0 & = D y_1 + B x_1 - Fz_1 \label{eq: component balance}\\
0 & = A_1 - \frac{B_1}{C_1 + T} - \log_{10} p_1^{sat} \label{eq: Antoine} \\
        0 & = py_1 - p_1^{sat}x_1 \gamma_1 \label{eq: Raoult} 
    \end{align}
\end{subequations}
Here, Equations~\eqref{eq: Mass Balance} and \eqref{eq: component balance} are the total mole and component balances, Equation~\eqref{eq: Antoine} is Antoine's equation~\citep{Towler2013ChemicalEngineeringDesign}, and Equation~\eqref{eq: Raoult} is modified Raoult's law. 
Model~\eqref{eq: flash model} assumes an ideal gas phase and includes the activity coefficient for a nonideal liquid phase. 
Conservation laws, such as the mole balances, are fundamental principles in engineering modeling, and thermodynamic models, such as Antoine's equation and Raoult's law, are standard in undergraduate chemical engineering curricula. Hence, Model~\eqref{eq: flash model} can be easily derived without extensive research or specialization. 
Table~\ref{tab: Flash Variables} lists all variables in the equation system, and further values are listed in Appendix~\ref{App: Problem and property data}. 
\begin{table}
\caption{Variables of the flash unit. Index 1 denotes the volatile component.}
\label{tab: Flash Variables}
\begin{tabularx}{\columnwidth}{@{}lXl@{}}
\toprule
Variable   & Description             & Unit                          \\ \midrule
$F$        & Feed                    & mol/s \\$D$        & Distillate              & mol/s \\$B$        & Bottom                  & mol/s \\$z_1$      & Mole fraction feed       & [-] \\ $y_1$      & Mole fraction distillate & [-] \\ $x_1$      & Mole fraction bottom     & [-] \\ $p$        & Pressure                & [bar] \\ $p^{sat}$  & Vapor pressure          & [bar] \\ $T$        & Temperature             & [K] \\ $\gamma_1$ & Activity coefficient    & [-] \\ \bottomrule
\end{tabularx}
\end{table}

Equation System~\eqref{eq: flash model} has three DoF. With the two decision variables $p$ and $T$, the activity coefficient $\gamma_1$ remains as the unknown variable. 
Thus, we aim for a model for the activity coefficient. 
Thermodynamic knowledge states that the thermodynamic equilibrium is fully defined by two of the values in the temperature-pressure-mole fraction triplet~\citep{Towler2013ChemicalEngineeringDesign}. 
Thus, we can write the activity coefficient as a function of the temperature $T$ and the mole fraction in the bottom $x_1$, which is typical for activity coefficient models. 
Furthermore, the activity coefficient must be positive, and therefore, we write the model to predict the natural log of $\gamma_1$, which is also a common modeling choice for activity coefficient models~\citep{Towler2013ChemicalEngineeringDesign}:
\begin{equation}\label{eq: ln gamma}
    \ln \gamma_1 = h(T, x_1)
\end{equation}
As before, we approximate the unknown function using a GP. 
With a zero prior of the GP, Equation~\eqref{eq: ln gamma} also sets the ideal mixture assumption with $\gamma_1=1$ as the constant prior to the hybrid model BO. 

As the oracle function for the activity coefficient $\gamma_1$ for water in the binary mixture, we use the Nonrandom Two-Liquid (NRTL) model~\citep{Towler2013ChemicalEngineeringDesign}. 
For details on the NRTL equations and parameter values, see Appendix~\ref{App: Problem and property data}. 

In a real-world case, the activity coefficient cannot be measured directly. 
Thus, we use the approach outlined in Section~\ref{ssec: Data for the GP}. 
In each trial of our \emph{in silico} experiment, we fix temperature $T$ and pressure $p$ and measure the concentration at the bottom $x_1$. 
Next, we run the simulation of Equation System~\eqref{eq: flash model} by fixing the variable $x_1$ to the measured value and solving the model with two DoF for the activity coefficient. 
We now have the inputs and labels available to train the GP. 
Hence, we can insert the GP into Model~\eqref{eq: flash model} and write the acquisition problem, e.g., the SAA-EI formulation in Equation~\eqref{eq: SAA EI}.

The bounds set for the input variables $T$ and $p$ allow for conditions in which a single phase is thermodynamically stable, so flash does not operate in the desired two-phase regime and model~\eqref{eq: flash model} is not valid. We consider these single-phase cases as infeasible, indicated by simulation results with $x_1>z_1$, i.e., higher mole fractions of the volatile component in the bottom compared to the feed, which is physically not possible for the non-azeotropic mixture. 
For the benchmark standard BO, we follow the concept by \cite{Wakabayashi2022BO_exp_fail} of including high values for the objective function as labels for the GP. 
We evaluate the objective function in Equation~\eqref{eq: flash Obj} with $y_1=0$ and the suggested values for temperature and pressure. 
This results in an almost flat response for the objective but maintains some gradients, as Equation~\eqref{eq: flash Obj} directly depends on the input variables. 
For the hybrid model BO case, the high objective approach by \cite{Wakabayashi2022BO_exp_fail} is not applicable as the GP does not predict the objective function, and the objective of the nonlinear hybrid is not directly proportional to the activity coefficient. 
For an infeasible trial, we set the label for the activity coefficient to one, i.e., $\ln(\gamma) = 0$. 
For the inputs, we have the temperature of the trial and the concentration at the bottom $x_1=0.5$. However, always setting the same value for $x$ will keep large sections of the input domain of the GP without training data and destabilize the GP training. 
Instead, we set a random value $x_1 =\tilde{x}_1\sim \mathcal{U}[0.5 , 1]$, where $\mathcal{U}$ is a uniform distribution. 
Thus, we obtain a training sample for the infeasible space of the system $([T, \tilde{x}], 0)$. 

\subsection{Convergence}\label{ssec: flash convergence}
We start the analysis of the flash optimization by looking at the convergence of the different methods. 
We initialize the BO loops with three initial points, sampled via Latin hypercube sampling in the temperature pressure space.
The hybrid BO experiments are run with 25 scenarios for $\xi_s$, and both BO loops use 100 multi-starts for the GP training and the acquisition problem optimization, respectively. 

\begin{figure}
    \centering
    \includegraphics[width=\columnwidth]{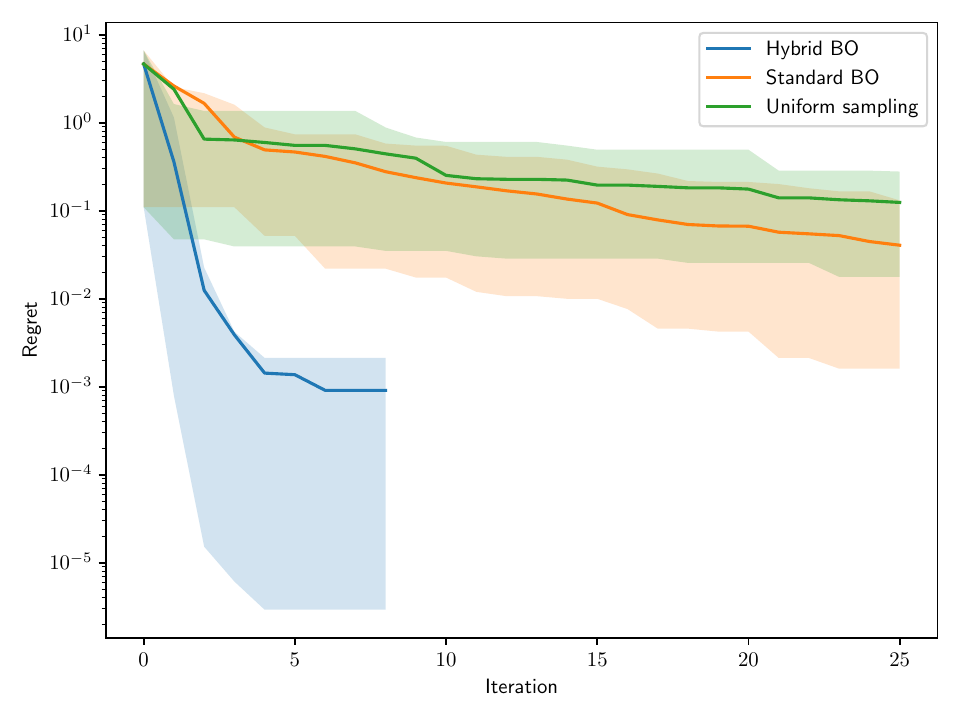}
    \caption{Log regret (see Equation~\eqref{eq: simple regret}) hybrid BO in comparison to standard BO with EI and random sampling.
    Confidence intervals from eight runs of the SAA-EI and 25 random and standard BO.}
    \label{fig: Flash regret}
\end{figure}
Figure~\ref{fig: Flash regret} shows the convergence over eight iterations of the hybrid model BO loop in comparison to 25 iterations of the benchmark standard BO and uniform random samples, again by showing the regret, see Equation~\eqref{eq: simple regret}. 
The regret is shown as the mean of the log regret over the BO runs (solid line) and the 5\% and 95\% quantiles (shaded regions).
The results show drastically improved performance of the hybrid model with SAA-EI over the standard BO EI approach and random sampling.
After just four iterations, the average hybrid model BO achieves objective values two orders of magnitude lower than the standard BO achieves after 25 iterations. 
The improved performance is a direct result of the additional knowledge written in the equations in model~\eqref{eq: flash model}. 
The model provides a structured relationship of the dependent variables and their effect on the objective function. 
Meanwhile, the standard BO is oblivious to the connection between the decision and dependent variables as well as the structural form of the objective, resulting in a higher number of trials for the optimization. 
Other than the convergence of the illustrative example, the regret values for the standard BO of the flash unit indicate a worse performance in absolute measures, as the absolute values are higher.
Hence, there is a significant improvement in both relative and absolute terms of the objective from formulating the hybrid model BO, which is further achieved within fewer iterations.

\subsection{Sampling Analysis}\label{ssec: flash sampling analysis}

\begin{figure}
    \centering
    \includegraphics[width=\columnwidth]{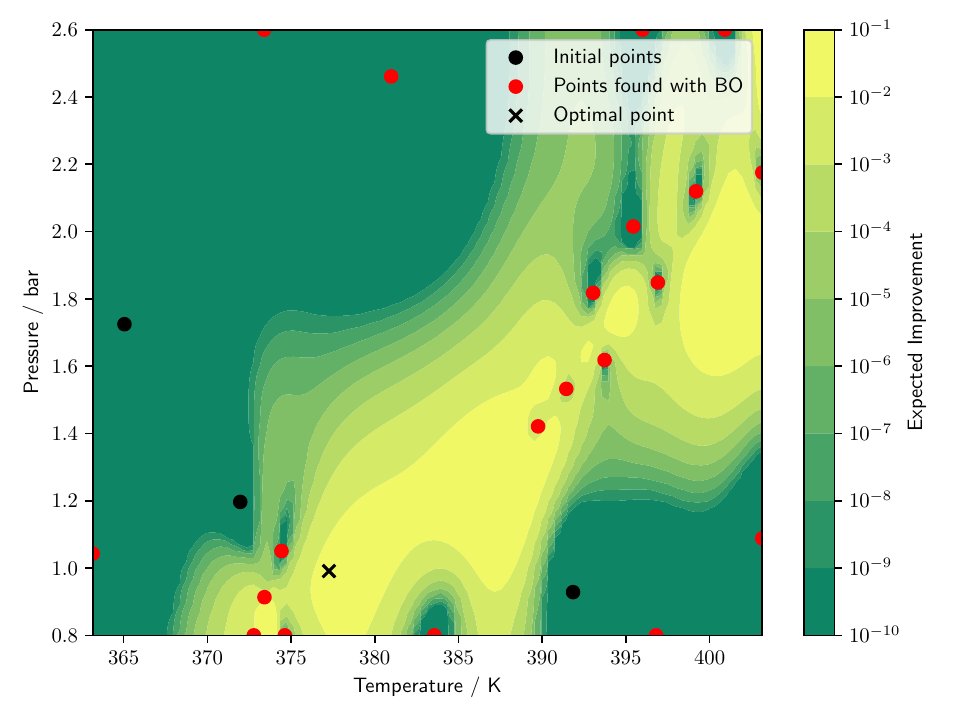}\\
    \includegraphics[width=\columnwidth]{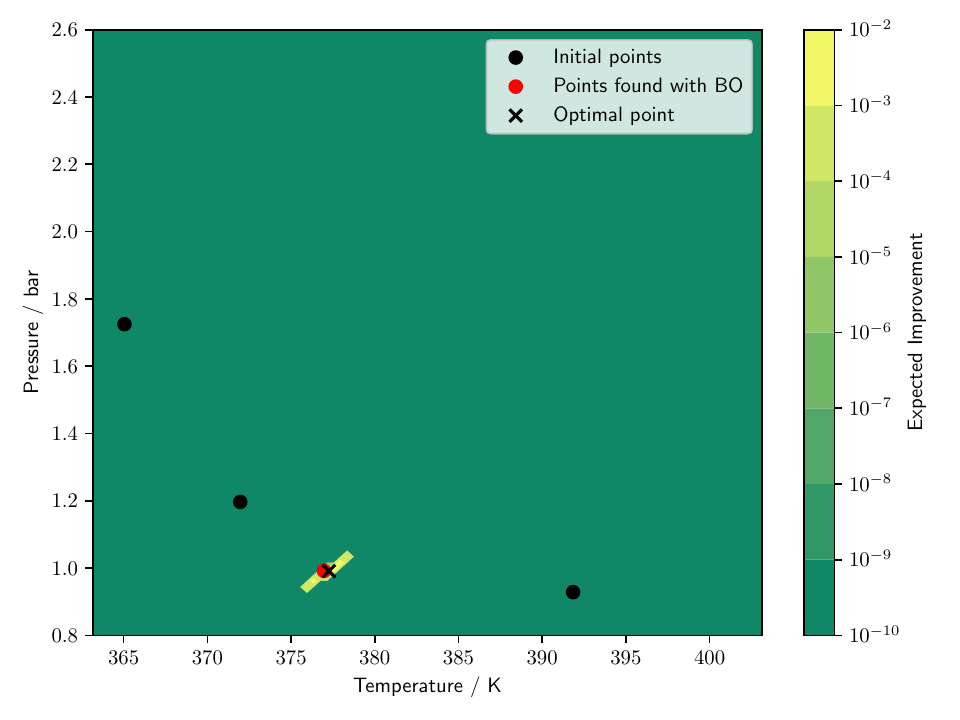}
    \caption{
        Comparison of the expected improvement (EI) acquisition functions in the Temperature - Pressure plane between the benchmark standard BO with EI (top) and the hybrid model BO using the SAA-EI function (bottom).
    }
    \label{fig: Flash Contour ACQ}
\end{figure}
We analyze the trial locations for a single run of a benchmark BO with an EI acquisition function in comparison to the hybrid model BO with the SAA-EI acquisition function. 
We use the same configuration for the BO loops as in Section~\ref{ssec: flash convergence}. 
Figure~\ref{fig: Flash Contour ACQ} shows the EI in standard BO (top) and SAA-EI in hybrid model BO (bottom) for the collected samples, respectively. 
The initial samples are shown in black, and the trials performed as part of the BO loops are shown in red. 
As observed in Section~\ref{ssec: flash convergence}, the hybrid model BO formulation converges significantly faster compared to the standard BO benchmark. 
The standard model BO shows 20 samples, which include a few infeasible trials.
For the BO with the hybrid model, we only show the first sampling location because the detected point is close to the global optimum, and further iterations do not improve the solution.
In Figure~\ref{fig: Flash Contour ACQ}, the EI based on the standard BO samples clearly identifies the region of the optimum, but also returns high EI values for the upper right region, and future sampling locations are likely to be far from the true optimum. 
Meanwhile, the SAA-EI from the hybrid model returns zero for the majority of the input variable space. 
Only a small region near the optimum yields a nonzero SAA-EI, with values two orders of magnitude smaller than the standard BO EI, and the remainder of the variable space returns zero SAA-EI.

\subsection{Discussion}
The analysis of flash unit optimization indicates that the hybrid BO model can yield significant benefits, specifically a reduction in the number of required trials. 
The comparison in Figure~\ref{fig: Flash Contour ACQ} shows that the hybrid model can rule out infeasible and suboptimal regions purely based on available physical knowledge, without sampling in these regions. 
By writing the hybrid model, major parts of the objective are defined by the mechanistic parts of the model, and the BO loop simply has to narrow down the solution based on very few trials. 

Figure~\ref{fig: Flash Contour ACQ} shows the approximation of the EI via 25 samples of the GP posterior as described in Problem~\eqref{prob: SAA stochastic program}, and the true EI of the hybrid model BO may include larger non-zero domains. For continuous EI formulations in standard BO, the acquisition function vanishes numerically, and \cite{ament2023unexpectedImprovement} have proposed logarithmic EI formulations, which can lead to further improvement. However, the analysis of the illustrative example in Section~\ref{ssec: illustrative sampling analysis} has shown that the SAA approximation yields a zero SAA-EI as a result of the discretization. 
Thus, the analysis of the flash unit study confirms that the SAA-EI approximation is limited in convergence to near-optimal points.

\section{Conclusion}\label{sec: conclusion}
The presented hybrid modeling approach for BO allows us to write simple mechanistic models and compensate for the lack of knowledge on the details with an acquisition function-based BO loop. 
The concept combines the best of the two worlds of mechanistic modeling and machine learning based BO, minimizing both the time for modeling and the cost of experiments. 
The developed Python-based modeling environment implements GPs in CasADi as constraints to NLPs and interfaces the resulting stochastic programs with nonlinear solvers such as IPOPT. 
For both the abstract example and the flash unit optimization, we see improved results from writing the BO acquisition problem in the proposed hybrid model format. 
Still, the formulation of the hybrid model BO loop presents a challenging optimization problem and requires at least fundamental modeling knowledge. Thus, the additional work needs to be justified by the expected savings in the experiments. 

The analysis of the sampling locations for both examples shows that for later iterations the SAA-EI acquisition function returns as constant zero for either all or the majority of the search space, despite the BO loop not being converged to the global optimum. This is a result from the SAA approximation, which limits the accuracy of the model and leads for failures in the optimization of the acquisition problems. 
Future work will investigate advanced solutions to the stochastic program, including continuous approximations of the acquisition function to allow the hybrid model BO to converge to the global optimum. 

Another open question lies in the consideration of infeasible regions, which could be either described by known inequality constraints or unknown infeasible regions, i.e., unknown inequality constraints. 
The flash unit optimization example shows infeasible trials, which were handled using simple established approaches. Future work will write known inequalities into the acquisition problem and incorporate learning of infeasible spaces. 
Further open questions include the handling of modeling errors, handling of noisy experiments, and validation with real experiments. 

In conclusion, the analysis points to significant advantages in terms of reduction of trials from writing hybrid models for BO. 
There are numerous possible applications in chemical engineering, biochemistry, and other disciplines dealing with complex and partially understood problems.

\section*{Acknowledgment}
This project was partly funded by the Werner Siemens Foundation within the WSS project of the century ``catalaix''.

\section*{Author Contribution}
E.C.: 
Conceptualization,
Formal analysis,
Investigation,
Methodology,
Supervision,
Visualization,
Writing – original draft

\noindent
L.K.: 
Investigation,
Data curation,
Software,
Visualization,
Writing – review \& editing

\noindent
O.S.: 
Investigation,
Software,
Visualization
Writing – review \& editing

\noindent
J.A.P.:
Conceptualization,
Software,
Methodology,
Writing – review \& editing

\noindent
A.M.:
Funding acquisition,
Writing – review \& editing

\appendix

\section{Lower Confidence Bound}\label{app: LCB}
One of the most popular and intuitive acquisition functions is the lower confidence bound (LCB) for minimization problems. 
The LCB describes the lower edge of the confidence region predicted by the GP. 
Hence, the general form of the LCB is given by~\citep{garnett_bayesoptbook_2023}:
\begin{equation}\label{eq: LCB general}
    \alpha_{\mathrm{LCB}}(\mathbf{u}) = \mu_\phi(\mathbf{u}) - \beta k_{\phi}(\mathbf{u}, \mathbf{u})^{0.5}
\end{equation}
Here, $\mu_\phi$ and $k_\phi$ are the posterior mean and covariance functions of the GP estimate of the unknown objective function $\phi$. The parameter $\beta$ is a scaling hyperparameter.

For the hybrid BO formulation, the LCB acquisition problem uses the statistical moments of the objective $f$. 
As there is no closed-form expression for the nonlinear case, we approximate $\mu_f$ and $\sigma_f$ via the SAA of the mean and the variance:
\begin{align}
    \hat{\mu}_f &\approx \frac{1}{S}\sum_{s=1}^S f(\mathbf{x}_s,\mathbf{y}_s, \mathbf{u}) \label{eq: SAA Objective mean}\\
    \hat{\sigma}^2_f &\approx \frac{1}{S-1}\sum_{s=1}^S \left(f(\mathbf{x}_s,\mathbf{y}_s, \mathbf{u}) - \hat{\mu}_f\right)^2  \label{eq: SAA Objective variance}
\end{align}
To write the LCB acquisition problem, we simply add Equation~\eqref{eq: SAA Objective mean} and \eqref{eq: SAA Objective variance} to Problem~\eqref{prob: SAA stochastic program} and use the SAA of the moments to formulate the objective:
\begin{equation}
    \underset{\mathbf{u}}{\min}~\hat{\mu}_f - \beta \hat{\sigma}_f
\end{equation}

\section{Smooth Sample Average Expected Improvement}\label{app: smooth SAA EI}
For cases where the nonsmooth SAA-EI function in Equation~\eqref{eq: SAA EI} does not work, we propose the following smooth approximation: 
\begin{equation}
    \min(0,f - f') =
    \frac{(f - f') - \sqrt{(f - f')^2+\varepsilon}}{2}
\end{equation}
The full objective of the EI acquisition problem then reads:
\begin{equation}\label{eq: smooth SAA EI}
    \underset{\mathbf{u}\in \mathcal{U}}{\min}~ 
    \frac{1}{2S}\sum_{s=1}^S 
    (f_s - f') - \sqrt{(f_s - f')^2+\varepsilon}
\end{equation}
We add a small $\varepsilon$ to the term under the square root to ensure numerical stability. 

Another alternative can be written using a minimization version of the soft plus function:
\begin{equation}
    \min(0,f - f')\approx-\ln\left(1+\exp(-(f - f'))\right)
\end{equation}
Here, the SAA approximation reads:
\begin{equation}
    \underset{\mathbf{u}\in \mathcal{U}}{\min}~ 
    \frac{1}{S}\sum_{s=1}^S 
    -\ln\left(1+\exp(-(f_s - f'))\right)
\end{equation}

\section{Flash Parameters and Property Data}\label{App: Problem and property data}

This Section contains the problem parameters and thermodynamic property data required for the flash optimization problem in Section~\ref{sec: flash unit}.

Table~\ref{tab: problem parameters} lists all the problem parameters of the flash case study in Section~\ref{sec: flash unit}.

\begin{table}
\centering
\caption{Parameters and fixed values for flash unit case study.}
\begin{tabularx}{\columnwidth}{lXr}
\toprule
Symbol & Description & Value \\
\midrule
$y_{1,set}$ & Purity specification of $y_1$ & $0.66$ \\ 
$p_{amb}$ & Ambient pressure & $1.01325 \, \mathrm{bar}$ \\ 
$F$ & Feed flow rate & $1.0 \, \mathrm{mol/s}$ \\
$D$ & Top stream flow rate & $0.3 \, \mathrm{mol/s}$ \\
$z_1$ & Mole fraction of H$_2$O in feed & $0.5$ \\
$\zeta_1$ & Cost factor purity violation of $y_1$ & $100 \, \mathrm{\$}$ \\ 
$\zeta_2$ & Cost factor temperature supply        & $0.01 \,  \mathrm{\$/K^2}$ \\
$\zeta_3$ & Cost factor pressure supply           & $0.5 \, \mathrm{\$/bar^2}$ \\
\bottomrule
\end{tabularx}
\label{tab: problem parameters}
\end{table}

The Antoine equation, as delineated in Equation~\eqref{eq: Antoine}, employs component-specific parameters $A_i$, $B_i$, and $C_i$ for the calculation of vapor pressure $p_i^{sat}$ for a component $i$. We take the Antoine parameters of water from \cite{Poling2020PropertiesOfGasesLiquidsa} and convert to pressure in $\mathrm{Pa}$ and temperature in $\mathrm{K}$.

The activity coefficient given by the NRTL model for water $\gamma_1$ in the binary mixture is~\citep{Pfennig2004ThermodnamikGemische,AspenProperties2024}:
\begin{equation}\label{eq: NRTL activity coefficient}
    \begin{aligned}
    \ln \gamma_1 &=  x_2^2 \left[ \tau_{21} \left( \frac{G_{21}}{x_1 + x_2 G_{21}} \right)^2 + \frac{\tau_{12}G_{12}}{(x_2 + x_1 G_{12})^2} \right]
    \end{aligned}
\end{equation}
The NRTL model for binary mixture involves the three variables $\tau_{ij}$, $G_{ij}$, and $\alpha_{ij}$ to describe the nonideal interactions between the components $i$ and $j$. The interaction parameters $G_{12}$ and $G_{21}$ are calculated as:
\begin{equation}\label{eq: NRTL G parameters}
    \begin{aligned}
    G_{12} &= \exp\left( -\alpha_{12} \tau_{12} \right) \\
    G_{21} &= \exp\left( -\alpha_{21} \tau_{21} \right)
    \end{aligned}
\end{equation}
The temperature-dependent interaction parameters $\tau_{12}$ and $\tau_{21}$ are expressed as:
\begin{equation}\label{eq: NRTL tau parameters}
    \begin{aligned}
    \tau_{12} &= a_{12} + \frac{b_{12}}{T} \\
    \tau_{21} &= a_{21} + \frac{b_{21}}{T} 
    \end{aligned}
\end{equation}
The factors $\alpha_{12}$ and $\alpha_{21}$ are:
\begin{equation}\label{eq: NRTL alpha parameters}
    \begin{aligned}
    \alpha_{12} &= c_{12} + d_{12} (T - 273.15) \\
    \alpha_{21} &= c_{21} + d_{21} (T - 273.15)
    \end{aligned}
\end{equation}
The NRTL parameters $a_{ij}$, $b_{ij}$, $c_{ij}$, and $d_{ij}$ are sourced from Aspen Properties~\citep{AspenProperties2024}. All the other model parameters are summarized in Table~\ref{tab: problem parameters}.

The NRTL parameters $a_{ij}$, $b_{ij}$, $c_{ij}$, and $d_{ij}$ for the binary mixture of water and acetic acid are taken from Aspen Properties~\citep{AspenProperties2024}.
Within Aspen Properties, the binary interaction data set \textit{NISTV110 NIST-HOC} is chosen. For the specified NRTL parameters, the input variable temperature $T$ is constrained within a validity range of $290.78\,\mathrm{K}$ at the lower bound to $504.83\,\mathrm{K}$ at the upper bound~\citep{AspenProperties2024}.

\bibliographystyle{apalike}
  \renewcommand{\refname}{Bibliography}  \bibliography{References.bib}

\begin{thebibliography}{}

\bibitem[Ament et~al., 2023]{ament2023unexpectedImprovement}
Ament, S., Daulton, S., Eriksson, D., Balandat, M., and Bakshy, E. (2023).
\newblock Unexpected improvements to expected improvement for bayesian optimization.
\newblock In Oh, A., Naumann, T., Globerson, A., Saenko, K., Hardt, M., and Levine, S., editors, {\em Advances in Neural Information Processing Systems}, volume~36, pages 20577--20612. Curran Associates, Inc.

\bibitem[Andersson et~al., 2018]{CasADi}
Andersson, J. A.~E., Gillis, J., Horn, G., Rawlings, J.~B., and Diehl, M. (2018).
\newblock Casadi: a software framework for nonlinear optimization and optimal control.
\newblock {\em Mathematical Programming Computation}, 11(1):1–36.

\bibitem[Aspen~Technology, 2024]{AspenProperties2024}
Aspen~Technology, I. (2024).
\newblock Aspen properties® v11.

\bibitem[Astudillo and Frazier, 2019]{astudillo2019_BO_composite}
Astudillo, R. and Frazier, P. (2019).
\newblock {B}ayesian optimization of composite functions.
\newblock In Chaudhuri, K. and Salakhutdinov, R., editors, {\em Proceedings of the 36th International Conference on Machine Learning}, volume~97 of {\em Proceedings of Machine Learning Research}, pages 354--363. PMLR.

\bibitem[Astudillo and Frazier, 2021]{astudillo2021bayesianFunctionNetworks}
Astudillo, R. and Frazier, P. (2021).
\newblock Bayesian optimization of function networks.
\newblock In Ranzato, M., Beygelzimer, A., Dauphin, Y., Liang, P., and Vaughan, J.~W., editors, {\em Advances in Neural Information Processing Systems}, volume~34, pages 14463--14475. Curran Associates, Inc.

\bibitem[Berkenkamp et~al., 2021]{Berkenkamp2021BOsafeRobotics}
Berkenkamp, F., Krause, A., and Schoellig, A.~P. (2021).
\newblock Bayesian optimization with safety constraints: safe and automatic parameter tuning in robotics.
\newblock {\em Machine Learning}, 112(10):3713–3747.

\bibitem[Biegler, 2010]{Biegler2010_NLP}
Biegler, L.~T. (2010).
\newblock {\em Nonlinear Programming: Concepts, Algorithms, and Applications to Chemical Processes}.
\newblock Society for Industrial and Applied Mathematics.

\bibitem[Buathong et~al., 2024]{Frazier2024BOFunNetworksPartialEval}
Buathong, P., Wan, J., Astudillo, R., Daulton, S., Balandat, M., and Frazier, P.~I. (2024).
\newblock {B}ayesian optimization of function networks with partial evaluations.
\newblock In Salakhutdinov, R., Kolter, Z., Heller, K., Weller, A., Oliver, N., Scarlett, J., and Berkenkamp, F., editors, {\em Proceedings of the 41st International Conference on Machine Learning}, volume 235 of {\em Proceedings of Machine Learning Research}, pages 4752--4784. PMLR.

\bibitem[Bynum et~al., 2021]{pyomo}
Bynum, M.~L., Hackebeil, G.~A., Hart, W.~E., Laird, C.~D., Nicholson, B.~L., Siirola, J.~D., Watson, J.-P., and Woodruff, D.~L. (2021).
\newblock {\em Pyomo — Optimization Modeling in Python}.
\newblock Springer International Publishing.

\bibitem[Ceccon et~al., 2022]{omlt}
Ceccon, F., Jalving, J., Haddad, J., Thebelt, A., Tsay, C., Laird, C.~D., and Misener, R. (2022).
\newblock Omlt: Optimization \& machine learning toolkit.
\newblock {\em Journal of Machine Learning Research}, 23(349):1--8.

\bibitem[Chitre et~al., 2023]{Lapkin2023pHbot}
Chitre, A., Cheng, J., Ahamed, S., Querimit, R. C.~M., Zhu, B., Wang, K., Wang, L., Hippalgaonkar, K., and Lapkin, A.~A. (2023).
\newblock phbot: Self‐driven robot for ph adjustment of viscous formulations via physics‐informed‐ml**.
\newblock {\em Chemistry–Methods}, 4(2).

\bibitem[Eugene et~al., 2023]{Dowling2023Bayesian_hybrid_models}
Eugene, E.~A., Jones, K.~D., Gao, X., Wang, J., and Dowling, A.~W. (2023).
\newblock Learning and optimization under epistemic uncertainty with bayesian hybrid models.
\newblock {\em Computers \& Chemical Engineering}, 179:108430.

\bibitem[Folch et~al., 2023]{Misner2023multi_fidelity_asychronous_batch}
Folch, J.~P., Lee, R.~M., Shafei, B., Walz, D., Tsay, C., van~der Wilk, M., and Misener, R. (2023).
\newblock Combining multi-fidelity modelling and asynchronous batch bayesian optimization.
\newblock {\em Computers \& Chemical Engineering}, 172:108194.

\bibitem[Gardner et~al., 2014]{Gardner2014BO_ineq}
Gardner, J., Kusner, M., Xiang, Z., Weinberger, K., and Cunningham, J. (2014).
\newblock Bayesian optimization with inequality constraints.
\newblock In Xing, E.~P. and Jebara, T., editors, {\em Proceedings of the 31st International Conference on Machine Learning}, volume~32 of {\em Proceedings of Machine Learning Research}, pages 937--945, Bejing, China. PMLR.

\bibitem[Garnett, 2023]{garnett_bayesoptbook_2023}
Garnett, R. (2023).
\newblock {\em {Bayesian Optimization}}.
\newblock Cambridge University Press.

\bibitem[González and Zavala, 2024]{Gonzlez2024BOIS}
González, L.~D. and Zavala, V.~M. (2024).
\newblock Bois: Bayesian optimization of interconnected systems.
\newblock {\em IFAC-PapersOnLine}, 58(14):446–451.

\bibitem[González and Zavala, 2025]{Gonzlez2025implementation_BOIS}
González, L.~D. and Zavala, V.~M. (2025).
\newblock Implementation of a bayesian optimization framework for interconnected systems.
\newblock {\em Industrial \& Engineering Chemistry Research}, 64(4):2168–2182.

\bibitem[Kingma and Welling, 2013]{Kingma2013VariationalBayesAndReparameterization}
Kingma, D.~P. and Welling, M. (2013).
\newblock Auto-encoding variational bayes.

\bibitem[Kudva and Paulson, 2026]{Paulson2026BONSAI}
Kudva, A. and Paulson, J.~A. (2026).
\newblock Bonsai: Structure-exploiting robust bayesian optimization for networked black-box systems under uncertainty.
\newblock {\em Computers \& Chemical Engineering}, 204:109393.

\bibitem[Le~Gratiet and Garnier, 2014]{LeGratiet2014recursivecokriging}
Le~Gratiet, L. and Garnier, J. (2014).
\newblock Recursive co-kriging model for design of computer experiments with multiple levels of fidelity.
\newblock {\em International Journal for Uncertainty Quantification}, 4(5):365–386.

\bibitem[Lima et~al., 2025]{Lima2025InnovationsProcessControl}
Lima, F.~V., Tian, Y., Durand, H.~E., Paulson, J.~A., and Biegler, L.~T. (2025).
\newblock Innovations in chemical process control: challenges and opportunities.
\newblock {\em Current Opinion in Chemical Engineering}, 48:101148.

\bibitem[Mainini et~al., 2025]{Mainini2025BenchmarkFunctions}
Mainini, L., Serani, A., Pehlivan-Solak, H., Di~Fiore, F., Rumpfkeil, M.~P., Minisci, E., Quagliarella, D., Yildiz, S., Ficini, S., Pellegrini, R., Thelen, A., Bryson, D., Nikbay, M., Diez, M., and Beran, P.~S. (2025).
\newblock Analytical benchmark problems and methodological framework for the assessment and comparison of multifidelity optimization methods.
\newblock {\em Archives of Computational Methods in Engineering}.

\bibitem[Paulson and Lu, 2022]{Paulson2022COBALT}
Paulson, J.~A. and Lu, C. (2022).
\newblock Cobalt: Constrained bayesian optimization of computationally expensive grey-box models exploiting derivative information.
\newblock {\em Computers \& Chemical Engineering}, 160:107700.

\bibitem[Paulson and Tsay, 2025]{PaulsonTsay2025BO_Opinion}
Paulson, J.~A. and Tsay, C. (2025).
\newblock Bayesian optimization as a flexible and efficient design framework for sustainable process systems.
\newblock {\em Current Opinion in Green and Sustainable Chemistry}, 51:100983.

\bibitem[Petsagkourakis et~al., 2022]{Petsagkourakis2022CC_Policy}
Petsagkourakis, P., Sandoval, I.~O., Bradford, E., Galvanin, F., Zhang, D., and Rio-Chanona, E. A.~d. (2022).
\newblock Chance constrained policy optimization for process control and optimization.
\newblock {\em Journal of Process Control}, 111:35–45.

\bibitem[Pfennig, 2004]{Pfennig2004ThermodnamikGemische}
Pfennig, A. (2004).
\newblock {\em Thermodynamik der Gemische}.
\newblock Springer Berlin Heidelberg.

\bibitem[Poling, 2020]{Poling2020PropertiesOfGasesLiquidsa}
Poling, B.~E. (2020).
\newblock {\em Properties of Gases and Liquids, Fifth Edition}.
\newblock McGraw-Hill's AccessEngineering. McGraw-Hill Education, New York, N.Y., fifth edition. edition.
\newblock Includes bibliographical references and index. - Description based on e-Publication PDF.

\bibitem[Rasmussen and Williams, 2005]{Rasmussen2005_GP4ML}
Rasmussen, C.~E. and Williams, C. K.~I. (2005).
\newblock {\em Gaussian Processes for Machine Learning}.
\newblock The MIT Press.

\bibitem[Reker et~al., 2020]{Reker2020BO_outperforms_Researchers}
Reker, D., Hoyt, E.~A., Bernardes, G.~J., and Rodrigues, T. (2020).
\newblock Adaptive optimization of chemical reactions with minimal experimental information.
\newblock {\em Cell Reports Physical Science}, 1(11):100247.

\bibitem[Ruszczyński and Shapiro, 2021]{Ruszczyski2021TwoStageProblems}
Ruszczyński, A. and Shapiro, A. (2021).
\newblock {\em Chapter 2: Two-Stage Problems}, page 21–52.
\newblock Society for Industrial and Applied Mathematics.

\bibitem[Schweidtmann et~al., 2021]{Schweidtmann2021DGO_GP}
Schweidtmann, A.~M., Bongartz, D., Grothe, D., Kerkenhoff, T., Lin, X., Najman, J., and Mitsos, A. (2021).
\newblock Deterministic global optimization with gaussian processes embedded.
\newblock {\em Mathematical Programming Computation}, 13(3):553–581.

\bibitem[Schweidtmann et~al., 2024]{Schweidtmann2024hybrid_modeling_Perspective}
Schweidtmann, A.~M., Zhang, D., and von Stosch, M. (2024).
\newblock A review and perspective on hybrid modeling methodologies.
\newblock {\em Digital Chemical Engineering}, 10:100136.

\bibitem[Shahriari et~al., 2016]{Shahriari2016Review_BO}
Shahriari, B., Swersky, K., Wang, Z., Adams, R.~P., and de~Freitas, N. (2016).
\newblock Taking the human out of the loop: A review of bayesian optimization.
\newblock {\em Proceedings of the IEEE}, 104(1):148–175.

\bibitem[Shapiro, 2009]{Shapiro2009SAA}
Shapiro, A. (2009).
\newblock {\em 5. Statistical Inference}, page 155–252.
\newblock Society for Industrial and Applied Mathematics.

\bibitem[Shapiro et~al., 2021]{Shapiro2021LecturesStochasticProgramming}
Shapiro, A., Dentcheva, D., and Ruszczynski, A. (2021).
\newblock {\em Lectures on Stochastic Programming: Modeling and Theory, Third Edition}.
\newblock Society for Industrial and Applied Mathematics.

\bibitem[Shields et~al., 2021]{Shields2021BO_reaction}
Shields, B.~J., Stevens, J., Li, J., Parasram, M., Damani, F., Alvarado, J. I.~M., Janey, J.~M., Adams, R.~P., and Doyle, A.~G. (2021).
\newblock Bayesian reaction optimization as a tool for chemical synthesis.
\newblock {\em Nature}, 590(7844):89–96.

\bibitem[Terayama et~al., 2021]{Terayama2021blackboxOpt}
Terayama, K., Sumita, M., Tamura, R., and Tsuda, K. (2021).
\newblock Black-box optimization for automated discovery.
\newblock {\em Accounts of Chemical Research}, 54(6):1334–1346.

\bibitem[T\"onnis et~al., 2025]{Cramer2025Stefans_Multifidelity_ESCAPE}
T\"onnis, S., Kaven, L.~F., and Cramer, E. (2025).
\newblock Comparison of multi-fidelity modelling methods for bayesian optimization.
\newblock In {\em Proceedings of the 35th European Symposium on Computer Aided Process Engineering (ESCAPE 35)}, volume~4 of {\em ESCAPE 35}, page 1294–1299. PSE Press.

\bibitem[Towler, 2013]{Towler2013ChemicalEngineeringDesign}
Towler, G. (2013).
\newblock {\em Chemical engineering design}.
\newblock Elsevier, Butterworth-Heinemann, Amsterdam, second edition edition.
\newblock Description based on publisher supplied metadata and other sources.

\bibitem[Virtanen et~al., 2020]{scipy2020pythonlibrary}
Virtanen, P., Gommers, R., Oliphant, T.~E., Haberland, M., Reddy, T., Cournapeau, D., Burovski, E., Peterson, P., Weckesser, W., Bright, J., {van der Walt}, S.~J., Brett, M., Wilson, J., Millman, K.~J., Mayorov, N., Nelson, A. R.~J., Jones, E., Kern, R., Larson, E., Carey, C.~J., Polat, {\.I}., Feng, Y., Moore, E.~W., {VanderPlas}, J., Laxalde, D., Perktold, J., Cimrman, R., Henriksen, I., Quintero, E.~A., Harris, C.~R., Archibald, A.~M., Ribeiro, A.~H., Pedregosa, F., {van Mulbregt}, P., and {SciPy 1.0 Contributors} (2020).
\newblock {{SciPy} 1.0: Fundamental Algorithms for Scientific Computing in Python}.
\newblock {\em Nature Methods}, 17:261--272.

\bibitem[von Stosch et~al., 2014]{vonStosch2014hybrid_modeling_PSE}
von Stosch, M., Oliveira, R., Peres, J., and Feyo~de Azevedo, S. (2014).
\newblock Hybrid semi-parametric modeling in process systems engineering: Past, present and future.
\newblock {\em Computers \& Chemical Engineering}, 60:86–101.

\bibitem[W\"{a}chter and Biegler, 2005]{ipopt}
W\"{a}chter, A. and Biegler, L.~T. (2005).
\newblock On the implementation of an interior-point filter line-search algorithm for large-scale nonlinear programming.
\newblock {\em Mathematical Programming}, 106(1):25–57.

\bibitem[Wakabayashi et~al., 2022]{Wakabayashi2022BO_exp_fail}
Wakabayashi, Y.~K., Otsuka, T., Krockenberger, Y., Sawada, H., Taniyasu, Y., and Yamamoto, H. (2022).
\newblock Bayesian optimization with experimental failure for high-throughput materials growth.
\newblock {\em NPJ Computational Materials}, 8(1).

\bibitem[Wang et~al., 2022]{Dowling2022Bayesian_hybrid_models_escape}
Wang, J., Eugene, E.~A., and Dowling, A.~W. (2022).
\newblock {\em Scalable Stochastic Programming with Bayesian Hybrid Models}, page 1309–1314.
\newblock Elsevier.

\end{thebibliography}

\end{document}